\RequirePackage{fix-cm}
\documentclass[twocolumn]{svjour3}          
\smartqed  

\usepackage{cite}
\usepackage[pdftex]{graphicx}
\usepackage{ragged2e}
\usepackage[tight,footnotesize]{subfigure}
\usepackage{rotating}
\usepackage{graphicx}
\usepackage{amsmath,amssymb} 
\usepackage{array}
\usepackage{multirow}
\usepackage{colortbl}
\usepackage{amsfonts}
\usepackage{pifont}
\usepackage{xspace}
\usepackage{etoolbox}
\usepackage{overpic}
\usepackage{color}
\usepackage{microtype}
\usepackage{placeins}

\newcommand{\figref}[1]{Fig. \ref{#1}}
\newcommand{\tabref}[1]{Tab. \ref{#1}}

\newcommand{\orcid}[1]{\href{https://orcid.org/#1}{\includegraphics[width=10pt]{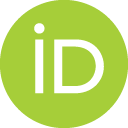}}}

\def\eg{\emph{e.g.}}

\def\etal{{\em et al}}

\usepackage{hyperref}
\hypersetup{breaklinks=true,citecolor=blue, colorlinks}

\graphicspath{{./Imgs/}}
\DeclareGraphicsExtensions{.pdf,.jpg,.png}

\usepackage{silence}
\journalname{Technical Report}

\begin{document}

\title{DriveMLM: Aligning Multi-Modal Large Language Models with \\ Behavioral Planning States for Autonomous Driving}

\titlerunning{E. Cui \etal, DriveMLM}        

\author{Erfei Cui$^{1,2*}$, Wenhai Wang$^{2,3*}$, Zhiqi Li$^{2,6*}$, Jiangwei Xie$^{4*}$, Haoming Zou$^{5*}$, Hanming Deng$^{4*}$, Gen Luo$^{2*}$, Lewei Lu$^{4}$, Xizhou Zhu$^{7}$, Jifeng Dai$^{7,8\dagger}$}

\authorrunning{E. Cui \etal} 

\institute{
$\dagger$Corresponding author: daijifeng@tsinghua.edu.cn \\
$^{7}$Department of Electronic Engineering, Tsinghua University, Beijing, 100084, China \\
$^{8}$Beijing National Research Center for Information Science and Technology, Beijing, 100084, China \\
Full list of author information is available at the end of the article $^{*}$ Equal contributors
}

\date{}

\maketitle

\begin{abstract}
Large language models (LLMs) have opened up new possibilities for intelligent agents, endowing them with human-like thinking and cognitive abilities. 
In this work, we delve into the potential of large language models (LLMs) in autonomous driving (AD). We introduce DriveMLM, an LLM-based AD framework that can perform close-loop autonomous driving in realistic simulators.
To this end, (1) we bridge the gap between the language decisions and the vehicle control commands by standardizing the decision states according to the off-the-shelf motion planning module. (2) We employ a multimodal LLM (MLLM) to model the behavior planning module of a module AD system, which uses driving rules, user commands, and inputs from various sensors (\eg, camera, lidar) as input and makes driving decisions and provide explanations; This model can plug-and-play in existing AD systems such as Autopilot  and Apollo for close-loop driving. (3) We design an effective data engine to collect a dataset that includes decision state and corresponding explanation annotation for model training and evaluation.
We conduct extensive experiments and show that replacing the decision-making modules of the Autopilot and Apollo with DriveMLM resulted in significant improvements of 3.2 and 4.7 points  on the CARLA Town05 Long respectively,
demonstrating the effectiveness of our model. We hope this work can serve as a baseline for autonomous driving with LLMs. 


\end{abstract}    
\section{Introduction}

\label{sec:intr}

\begin{figure}[t]
  \centering
  \subfigure[Rule-Based Autonomous Driving System~\cite{apollo_auto}]{
    \includegraphics[width=\linewidth]{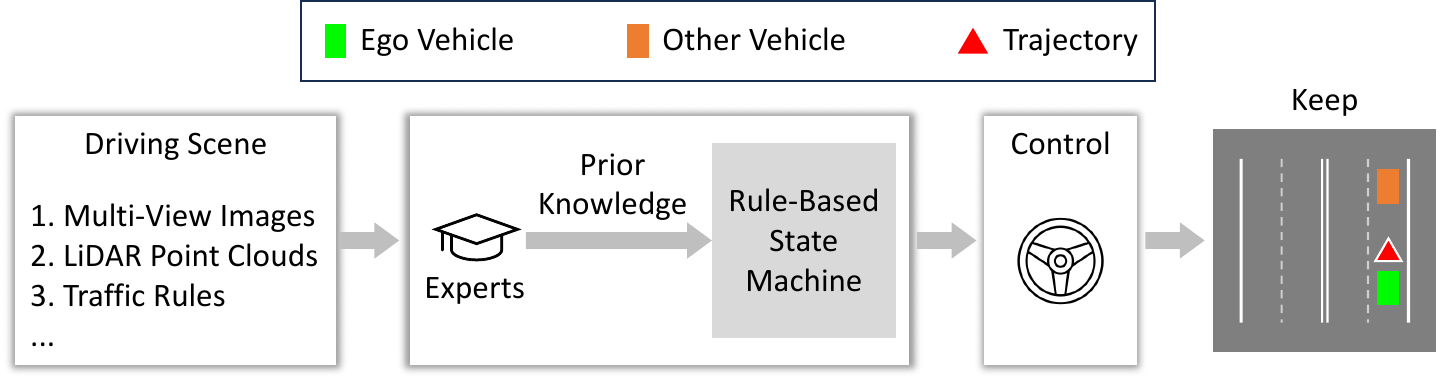}
    \label{fig:1a}
  }\\[-0.5ex]
  \subfigure[End-to-End Autonomous Driving System~\cite{hu2023planning,jia2023think,shao2023safety}]{
    \includegraphics[width=\linewidth]{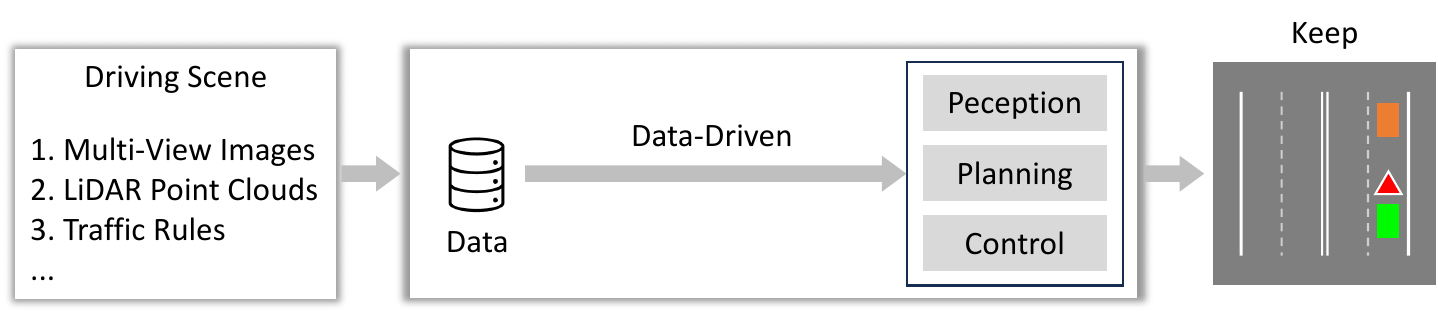}
    \label{fig:1b}
  }\\[-0.5ex]
  \subfigure[Autonomous Driving System with Large Language Model (Ours)]{
    \includegraphics[width=\linewidth]{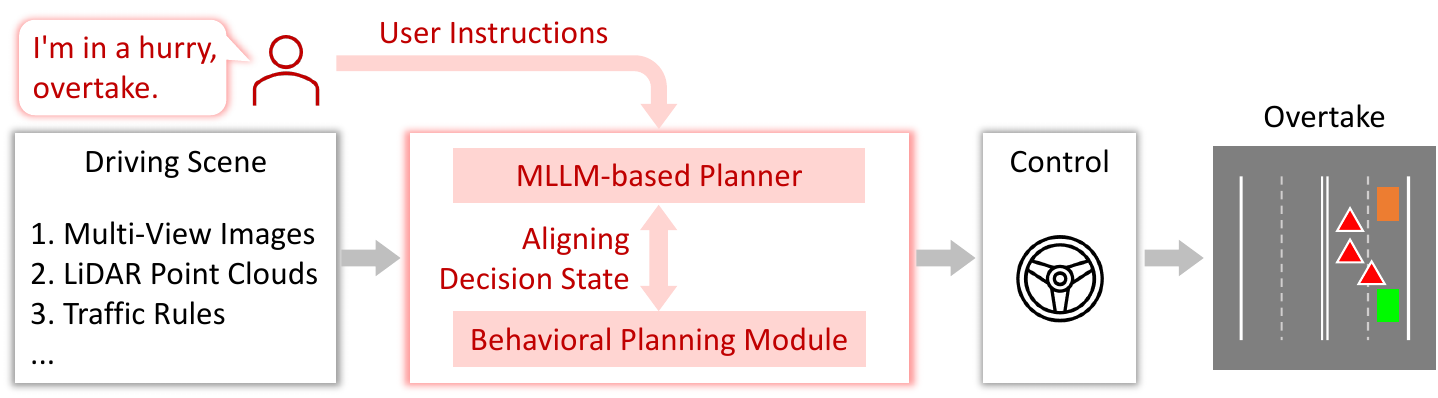}
    \label{fig:1c}
  }

  \caption{\textbf{Comparison of different AD systems.} (a) is rule-based with manually defined rules, (b) is data-driven but lacks diversity in training data, and (c) integrates LLM capabilities with aligned decision states for closed-loop planning.}
  \label{fig:moti}
\end{figure}

Autonomous driving (AD) has undergone significant advancements in recent years, evolving from traditional rule-based systems, which rely on a predefined set of rules informed by prior knowledge (see \figref{fig:1a}), to data-driven, end-to-end systems, as demonstrated in \figref{fig:1b}.
Despite their advancements, existing end-to-end models are limited to learning from autonomous driving scenario data, lacking sufficient understanding of the real world, hindering the models' ability to handle corner-case scenarios or those requiring complex reasoning. In contrast to traditional AD systems, Large language models (LLMs) trained with web-scale text corpus, are equipped with extensive world knowledge, robust logical reasoning, and advanced cognitive capabilities. These features position them as potential planners in AD systems, providing a human-like approach to autonomous driving.

Some recent studies \cite{drivelm2023, xu2023drivegpt4, wen2023dilu, liu2023mtd, hong2023metagpt, chen2023driving, sha2023languagempc} have been made to integrate LLMs into AD systems, focusing on generating language-based decisions in response to driving scenarios. 
However, these approaches have limitations when it comes to performing closed-loop driving in real-world environments or realistic simulators. This is because the outputs of LLMs are mainly linguistic and conceptual, which cannot be used for vehicle control.
In traditional modular AD systems~\cite{apollo_auto, autoware, fontana2021openpilot}, the gap between high-level strategic goals and low-level operational actions is connected by a behavioral planning module, whose decision states can be easily transformed into vehicle control signals by follow-up motion planning and control.
This motivates us to \emph{align the LLM with the decision state of the behavioral planning module, and further design an LLM-based close-loop AD system that can run on real-world environments or realistic simulators by using the aligned LLM for behavioral planning}.

Based on this point, we propose DriveMLM, an LLM-based AD framework that can perform close-loop autonomous driving in realistic simulators.  
To achieve this, we have three key designs: (1) We investigate the decision states of the behavioral planning module of the well-developed Apollo system~\cite{apollo_auto}, and transform them into forms that can be easily processed by LLMs. (2) We develop a multi-modal LLM (MLLM) planner that can accept the current multi-modal inputs including multi-view images, LiDAR point clouds, traffic rules, system messages, and user instructions, and predict the decision state; (3) To obtain enough training data for behavioral planning state alignment, we manually collect 280 hours of driving data on CARLA~\cite{dosovitskiy2017carla}, and convert them into decision state and corresponding explanation annotations by an efficient data engine. 
With these designs, we can obtain an MLLM planner that can make decisions based on the driving scenes and user requirements, and its decisions can be easily converted into vehicle control signals for closed-loop driving.

Our work offers several advantages, including:
(1) \textbf{Plug-and-Play.} Our MLLM planner's alignment with decision states allows seamless integration with existing modular autonomous driving (AD) systems like Apollo~\cite{apollo_auto} and AutoPilot \cite{tesla_autopilot}. This integration enables closed-loop driving without significant alterations or adjustments to the existing systems.
(2) \textbf{Interactive.} Leveraging the common knowledge and logical reasoning abilities of LLMs and taking language instructions as input, our model can comprehend high-level system messages, such as  basic driving logic and complex user instructions. This makes our model more flexible and adaptable to diverse driving scenarios and corner cases.

In summary, our main contributions are three folds:

(1) We propose an LLM-based AD framework that bridges the gap between LLM and closed-loop driving by aligning the output of LLMs with
the decision states of behavioral planning modules.

(2) To implement this framework, we tailor a set of decision states with forms that can be easily processed by LLMs, design an MLLM planner for decision prediction. In addition, we also develop a data engine that can effectively generate decision states and the corresponding explanation annotation for model training and evaluation.

\begin{figure}[t]
  \includegraphics[width=1.0\linewidth]{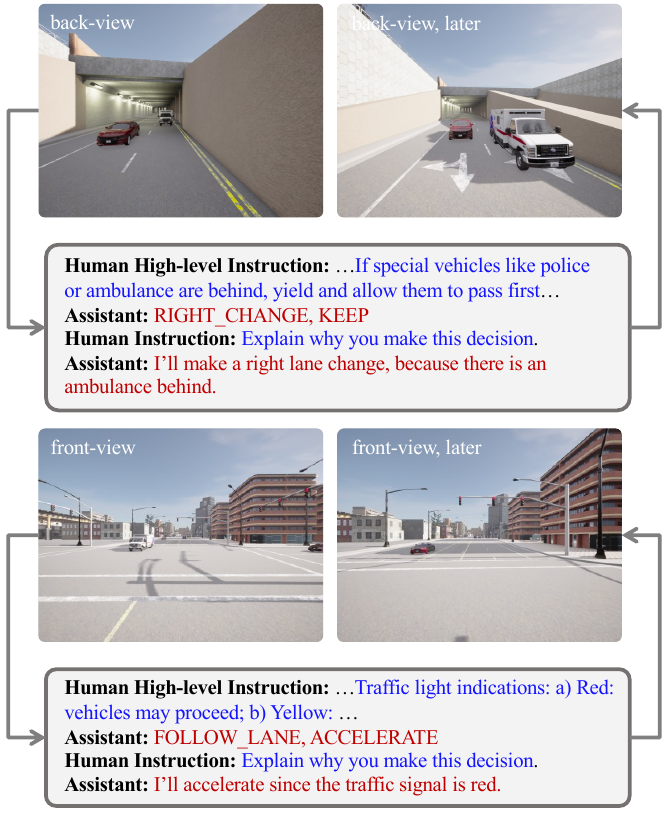}
  \caption{\textbf{Decision preferences altered by customized language instructions.} \textbf{TOP:} DriveMLM is instructed to yield when an emergency vehicle emerges from behind, leading to a lane change. \textbf{BOTTOM:} DriveMLM is instructed to proceed through a red light, causing a deviation from common traffic rules. In these scenarios, the driving system is influenced by modified driving preferences, resulting in unconventional control decisions.}
  \label{fig:vis_intro}   
\end{figure}

(3) To validate the effectiveness of our method, we not only evaluate our method on the closed-loop driving metrics including driving score (DS) and miles per intervention (MPI), but also use understanding metrics including accuracy, F1-measure for decision state, BLEU-4, CIDEr and METEOR for decision explanation to evaluate the driving understanding capability of our model. Notably, our method achieves 76.1 DS, 0.955 MPI results on CARLA Town05 Long, which is 4.7 points, 1.25 times better than Apollo~\cite{apollo_auto}. Moreover, our method demonstrates strong decision generalization capabilities when there are special requirements such as language instructions to hail an ambulance or obey traffic rules, as shown in \figref{fig:vis_intro}.

\section{Related Work}

\subsection{Multi-modal Large Language Models}

The swift evolution of Large Language Models (LLMs)~\cite{radford2018improving, radford2019language, brown2020language, ouyang2022training, gpt4} has recently given rise to the emergence of multi-modal LLMs (MLLMs)~\cite{alayrac2022flamingo, huang2023language, liu2023visual, liu2023improved, zhu2023minigpt, instructblip, ye2023mplug, chen2023shikra, wang2023visionllm, peng2023kosmos, lai2023lisa, bai2023qwen, gao2023llama, li2023videochat, zhang2023video, wu2023next, junqing2023never, zhang2023gpt4roi}, which augment language models with the capacity to analyze and comprehend information from diverse modalities.
Prominent instances of such advancements include GPT-4~\cite{gpt4}, Flamingo~\cite{alayrac2022flamingo}, KOSMOS-1~\cite{huang2023language}, LLaVA series \cite{liu2023visual, liu2023improved}, and MiniGPT-4 \cite{zhu2023minigpt}, as well as InstructBLIP~\cite{instructblip}. These models have integrated visual instruction tuning methodologies to enhance the MLLMs' ability to adhere to prescribed instructions.

Furthermore, mPLUG-DocOwl~\cite{ye2023mplug} has broadened the document comprehension capabilities of MLLMs by incorporating digital document datasets. Concurrently, Shikra~\cite{chen2023shikra}, VisionLLM~\cite{wang2023visionllm}, KOSMOS-2~\cite{peng2023kosmos}, LISA~\cite{lai2023lisa}, and Qwen-VL \cite{bai2023qwen} have augmented MLLMs with visual grounding capabilities, empowering them to detect or segment objects in accordance with user prompts. LLaVA-NeXt~\cite{liu2023visual} and InternVL-2~\cite{chen2024internvl,chen2024far}
leavage dynamic image splitting to handle high-resolution images.
Eagle-2~\cite{li2024eagle2fasterinferencelanguage} empolies different vision encoders including both SigLIP~\cite{zhai2023sigmoid}and ConvNeXt~\cite{liu2022convneXt} that enhances the visual features. 
The introduction of VideoChat~\cite{li2023videochat} and VideoLLaMA~\cite{zhang2023video} has ushered in the integration of video processing capabilities into LLMs. Additionally, NExT-GPT~\cite{wu2023next} has introduced a modality-switching instruction tuning technique for multi-modal prompt tuning, facilitating the handling of inputs and outputs in any combination of text, images, videos, and audio.
ASM~\cite{junqing2023never} and GPT4RoI~\cite{zhang2023gpt4roi} introduce region-level recognition and understanding capability into LLMs. These endeavors demonstrate the effectiveness and generalizability of LLMs, establishing a foundation for open-world tasks.

\subsection{Intelligent Agents with Large Language Models}

A burgeoning application of LLMs is their role in facilitating interaction and communication among intelligent agents (e.g., robots, virtual assistants, or game characters) and various entities, including humans, the environment, or even the intelligent agents themselves. Several API-based methods, including Visual ChatGPT~\cite{wu2023visual}, MM-REACT~\cite{yang2023mm}, HuggingGPT~\cite{shen2023hugginggpt}, InternGPT~\cite{liu2023internchat}, ViperGPT~\cite{suris2023vipergpt}, ControlLLM~\cite{liu2023controlllm}, and GPT4Tool~\cite{yang2023gpt4tools} have attempted to integrate diverse modal APIs with LLMs to accomplish complex tasks in the open world, such as image editing, video processing, and audio synthesis. These methods allow language models to perform complex real-world tasks by following natural language instructions. In parallel, alternative research initiatives, such as Camel~\cite{li2023camel}, AutoGPT~\cite{yang2023auto}, MetaGPT~\cite{hong2023metagpt} and Smallville~\cite{park2023generative}, investigate the utility of LLMs in the context of role-playing conversations or communication games.  Additionally, within the domain of embodied AI, works such as PaLM-E~\cite{driess2023palm}, EmbodiedGPT~\cite{mu2023embodiedgpt}, and the RT series~\cite{brohan2022rt, brohan2023rt, padalkar2023open} leverage LLMs to generate natural language actions, thereby controlling embodied agents proficient in executing navigation, manipulation, and interaction tasks within real or 3D environments. These works demonstrate the notable advancements achieved by LLMs in the realm of intelligent agent control.

\subsection{Autonomous Driving Models}

The development of autonomous driving (AD) models has accelerated rapidly in recent years, giving rise to many disruptive and groundbreaking technologies. Notably, the open-source frameworks, such as Apollo~\cite{apollo_auto} and Autoware~\cite{autoware}, have played pivotal roles by furnishing robust tools and resources, thereby facilitating the development of autonomous driving technology and contributing to its widespread adoption and progression.
In terms of AD perception, BEV (Bird’s Eye View)~\cite{li2022bevformer, yang2023bevformer, liang2022bevfusion, singh2023surround} and Occupancy Network~\cite{tong2023scene, shi2023grid, li2023fb} have become essential components of autonomous vehicles, helping them better understand the surrounding environment and make corresponding decisions.

The decision-making process in conventional autonomous driving systems typically relies on finite state machines~\cite{chen2019autonomous}. These systems often require the manual creation of numerous rules to determine the states and conditions for transitioning between them. However, considering the ever-changing nature of the world, this is usually laborious to design rules to cover all the scenarios for the real world.
In recent years, end-to-end autonomous driving models have also made remarkable progress, such as UniAD~\cite{hu2023planning}, which adopts a novel end-to-end approach, directly integrating perception, prediction, and planning, avoiding information loss and efficiency issues in the traditional modular design method. 

Recently, open-sourced simulators~\cite{dosovitskiy2017carla, vinitsky2022nocturne, zhou2020smarts} have been proposed to bridge the gap between model prediction and closed-loop control. Among them, CARLA~\cite{dosovitskiy2017carla}, featuring comprehensive sensor simulations and realistic environments, is the most widely used benchmark for evaluating closed-loop performance by many state-of-the-art methods~\cite{jia2023think, shao2023reasonnet, shao2023safety, jiang2023vad, chitta2022transfuser, chen2020learning, chen2021learning, chen2022learning}.

Recent works~\cite{drivelm2023,xu2023drivegpt4,mao2023gpt,wen2023dilu,liu2023mtd,chen2023driving,sha2023languagempc} changes our perception by introducing LLM for driving planning, opening up a new direction for the autonomous driving field.As early explorations, some~\cite{wen2023dilu,sha2023languagempc} use ChatGPT and GPT-4 to predict driving decisions. 
Following works fine-tune LLM models to predict driving signal~\cite{chen2023driving}, trajectory~\cite{mao2023gpt} or designed decision space~\cite{liu2023mtd}, conditioned only on language as input. 
DriveGPT4~\cite{xu2023drivegpt4} finetunes Multimodal LLM to 
predict control signal. 
However, DriveGPT4 is constrained by the input from a monocular camera, limiting its ability to construct comprehensive scene information.

All LLM-based works above are not evaluated on realistic simulators in closed-loop driving, because either linguistic decisions of LLMs are hard to transform to actually reliable control signals, or the direct prediction of control signal by LLM remains a large gap to real-time closed-loop driving.
\section{Proposed Method} 

\label{sec:method}

\subsection{System Overview}

\begin{figure*}[t]
  \includegraphics[width=1.0\linewidth]{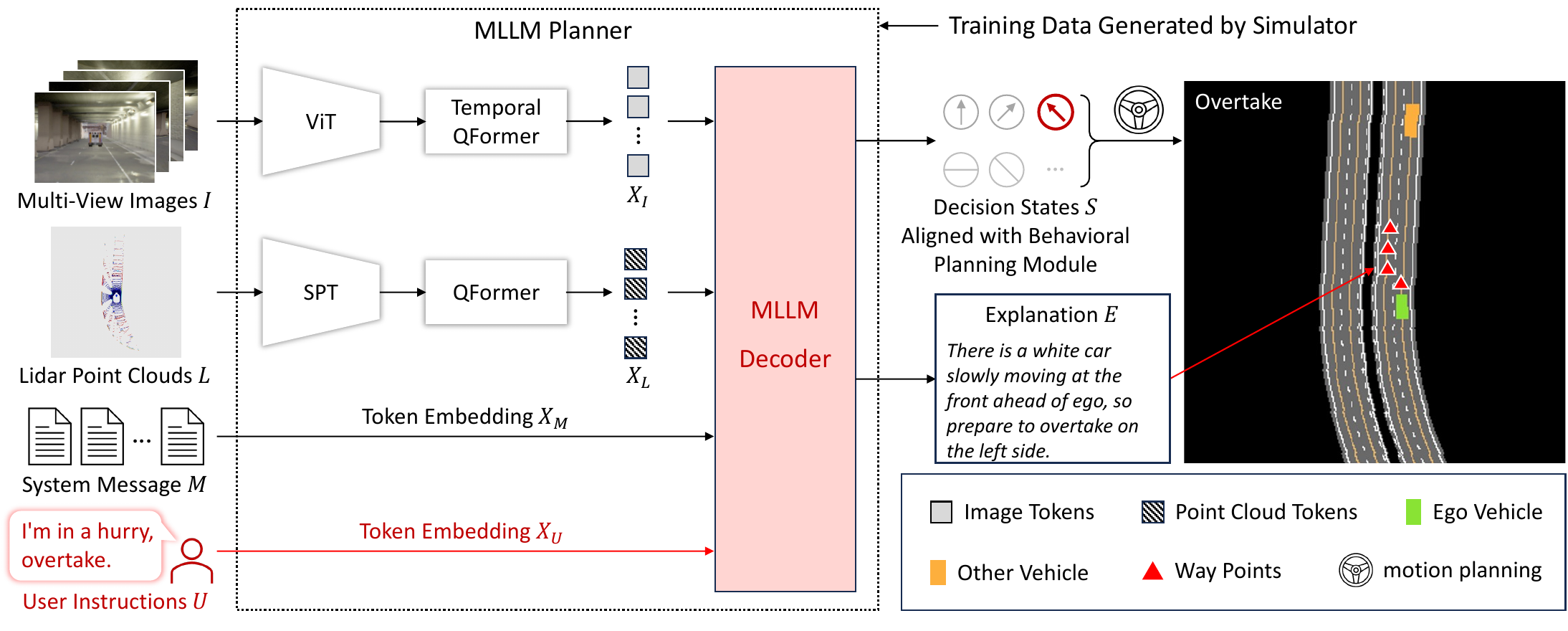}
  \caption{\textbf{DriveMLM framework} consists of three parts: (1) behavioral planning states alignment, which aligns linguistic output to executable decisions for vehicle control. (2) MLLM planner, consisting of multi-modal tokenizer and MLLM decoder. It transforms multi-modality sensor input $I,L$, system message $M$, and user instructions $U$ to drive explanation and aligned decisions. (3) Efficient data collection strategy, which generates rich driving explanations and aligned decisions at low cost. SST stands for single-stride sparse transformer.}
  \label{fig:main}       
\end{figure*}

The DriveMLM framework integrates the world knowledge and reasoning capabilities of large language models (LLMs) into an autonomous driving (AD) system, achieving closed-loop driving in realistic simulators. As illustrated in \figref{fig:main}, this framework has three key designs:
(1) \emph{Behavioral Planning States Alignment.} This part aligns LLM's linguistic decision outputs with the behavioral planning module of a well-established modular AD system like autopilot \cite{tesla_autopilot}. In this way, the output of LLM can be easily transformed into vehicle control signals.
(2) \emph{MLLM Planner.} It is a combination of a multi-modal tokenizer and a multi-modal LLM (MLLM) decoder. The multi-modal tokenizer transforms diverse inputs like multi-view images, LiDAR, traffic rules, and user requirements into unified tokens, and the MLLM decoder makes decisions based on the unified tokens. (3) \emph{Efficient Data Collection Strategy.} It introduces a tailored data collection method for LLM-based autonomous driving, ensuring a comprehensive dataset encompassing decision states, decision explanations, and user commands.

During inference, the DriveMLM framework leverages multi-modal data to make driving decisions. These data include: multi-view images $I\!\in\!\mathbb{R}^{T \times N_I \times H \times W \times 3}$, where $T$ denotes the time length, $N_I$ indicates the number of views, and $H$ and $W$ denotes the height and width of images. 
The point clouds $L\!\in\!\mathbb{R}^{K \times 4}$ from LiDAR point clouds, with $K$ representing the number of points. System message is denoted by $M\!\in\!\mathbb{R}^{N_M}$, and $N_M$ represents the number of system message tokens. The system message is the gathering of task definition, traffic rules, and decision state definition. User instructions $U\!\in\!\mathbb{R}^{N_U}$, where $N_U$ stands for the number of user instruction tokens.
These inputs undergo tokenization through a multi-modal tokenizer, resulting in: $X_I\!\in\!\mathbb{R}^{N_I \times N_Q \times D}$, $X_L\!\in\!\mathbb{R}^{1 \times D}$, 
$X_M\!\in\!\mathbb{R}^{N_M \times D}$, 
$X_U\!\in\!\mathbb{R}^{N_U \times D}$, 
which represent the tokens embedding of multi-view images, LiDAR point clouds, traffic rules, and user instructions, respectively.
Here, $N_Q$ denotes the number of output image token which is decided by the number of features in the last Transformer layer of the pre-trained CLIP visual encoder, and each token embedding is with $D$ dimension.
Next, these tokens are inputted into the MLLM decoder, which generates the decision state token $S$ along with a corresponding explanation $E$. 
Finally, the decision state $S$  is inputted into a motion planning and control module. This module computes the final trajectory for vehicle control.

\subsection{Behavioral Planning States Alignment}

Transforming the linguistic choices of Large Language Models (LLMs) into actionable control signals is crucial for vehicle control. To achieve this, we align the LLM's outputs with the decision stages of the behavioral planning module in the CARLA AutoPolot System.

We divide the decision-making process into two categories: speed  and path decisions.
Specifically, the speed decision states contain [KEEP, ACCELERATE, DECELERATE, STOP], while the path decision states include [FOLLOW, LEFT\_CHANGE, RIGHT\_CHANGE, LEFT\_BORROW, RIGHT\_BORROW]. 

\begin{figure}[t]
  \includegraphics[width=1.0\linewidth]{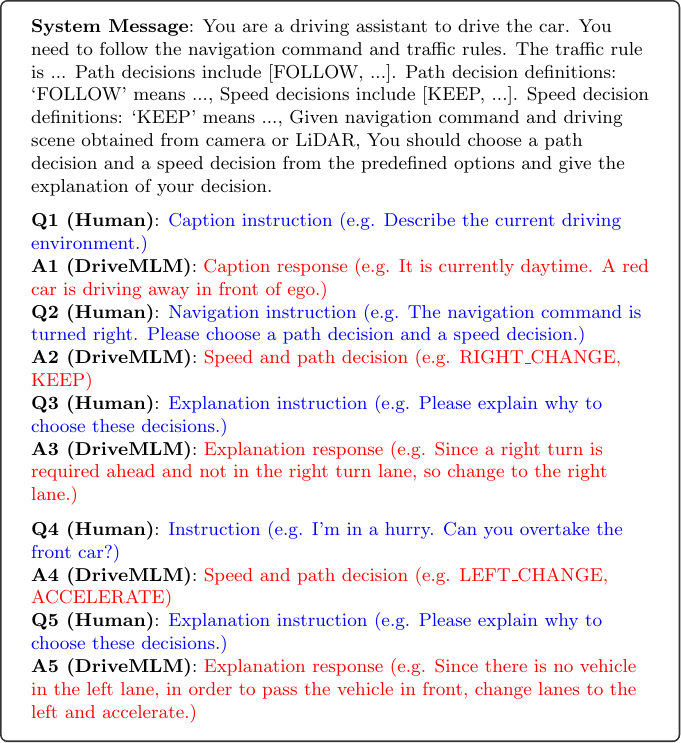}
  \caption{\textbf{Examples of system message and interaction between user and DriveMLM system.} The system message includes the description of the driving task, the traffic rules, and the definition of decision states. Given driving scenes such as images and user prompts, the driving system can infer the image caption, path, and speed decision, and additional explanation.}
  \label{fig:chat_system}       
\end{figure}

To enable a language model to make precise predictions among these states, we established a comprehensive link between linguistic descriptions and decision states, as illustrated in System Massage of \figref{fig:chat_system}. This correlation is used as a part of the system message and is integrated into the MLLM planner. As a result, once the LLM describes certain situations, the prediction will converge into a clear decision within the decision space. At each time, one-speed decision and one path decision are mutually inferred and sent to the motion planning framework. 

\subsection{MLLM Planner}

The MLLM planner of DriveMLM consists of two components: the multi-modal tokenizer and the MLLM decoder. The two components collaborate closely, handling a variety of inputs to accurately determine driving decisions and provide explanations for these decisions.

\noindent\textbf{Multi-Modal Tokenizer.} This tokenizer is engineered to handle three different forms of input efficiently. 

(1) For temporal multi-view images: We use a temporal cross-attention model to process multi-view images from timestamp $-T$ to 0 (current timestamp). First, images from various perspectives and times are processed by the CLIP vision encoder to obtain image features.
Next, these image features undergo a nonlinear transformation through a projector to be mapped to the same feature dimension as the text tokens. Finally, through the cross-attention layer, the current time's image is used as the query, while the historical images are first flattened into tokens, and used as key and value inputs to the multi-head cross-attention layer to obtain the fused features.

(2) For LiDAR data, it is necessary to align LiDAR features with text features into the same feature space, which enables the model to understand and process data from different modalities in a unified manner. However, text descriptions of scenarios are difficult to obtain due to the limitation of related datasets, we used the image domain as an intermediary for connecting text and point cloud. An image-lidar CLIP model is applied here, where the image encoder is the image encoder of ViT-L/14~\cite{radford2021_clip}, and we freeze all its parameters. For the LiDAR encoder, we utilized the single-stride sparse transformer (SST)\cite{fan2022embracing}, which is randomly initialized. During the training process, an image is sent to the frozen image encoder to produce the target feature $X_I$, while the LiDAR feature is extracted by the SST, 
 with the input of corresponding LiDAR data $X_L$. The LiDAR encoder is trained to maximize the similarity between $X_L$ and $X_I$ with a cosine similarity loss, for mapping the LiDAR features into image feature space. After obtaining the LiDAR feature $X_L$, we project it linearly to a LiDAR token $H_L$, which has the same feature dimensionality as the language tokens.  We applied data pre-processing during training. Every LiDAR data point corresponds to several views of images, and we regard each of them as a corresponding pair. The point cloud is also transformed into the camera coordinate system, and we project the 3D points into 2D image space and drop all the points outside the boundary of images, which are invisible. During test, we doesn't drop any points because there are multiple views for a timestep.

(3) For system messages and user instructions, we simply treat them as normal text data and use a token embedding layer of LLM to extract their embedding, $X_M\!\in\!\mathbb{R}^{N_M \times D}$, $X_U\!\in\!\mathbb{R}^{N_U \times D}$.

\noindent\textbf{MLLM Decoder.} The decoder is the core that translates the tokenized inputs into decision states and decision explanations. To this end, we design a system message template for LLM-based AD, which is shown in \figref{fig:chat_system}.
We see that the system messages contain a description of the AD tasks, traffic rules, the definition of decision states, and placeholders indicating where each modality's information is incorporated. This approach ensures that inputs from various modalities and sources are seamlessly integrated.

The output is formatted to provide decision states (see the Q2 of \figref{fig:chat_system}) and an explanation of the decisions (see the Q3 of \figref{fig:chat_system}), offering transparency and clarity in the decision-making process. Regarding the supervision methods, our framework uses cross-entropy loss with the next token prediction, following common practices. In this way, the MLLM planner can perform detailed understanding and processing of data from different sensors and sources, and transform it into appropriate decisions and explanations.

\subsection{Efficient Data Engine}

We propose a data generation pipeline that can create decision states and explanation annotations from various scenarios in the CARLA simulators. This pipeline can address the limitations of existing driving data, which lack decision states and detailed explanations for training LLM-based AD systems.Our pipeline consists of two main components: data collection and data annotation.

The data collection is designed to improve decision variety while staying realistic. 
First, various challenging scenarios are constructed in the simulator. Complex driving behaviors are required to safely drive through. 
Then, experts, either experienced human drivers or agents, are asked to safely drive through these scenarios triggered at one of its many passable locations. Notably, interaction data is generated when the expert randomly raises driving demand and drives accordingly. 
Once the expert drives safely to the destination, the data is recorded. 

The data annotation mainly focuses on decision and explanation. First, speed and path decision states are automatically annotated based on experts' driving trajectories by using hand-crafted rules. Second, explanation annotations are first generated based on the scenario, dynamically defined by current elements nearby. Third, the generated explanation annotations are refined by human annotators, and their variety is expanded by GPT-3.5. In addition, the interaction content is also refined by human annotators, including cases that are both executing or rejecting human requests.
In this way, we avoid the costly frame-by-frame decision state annotation, as well as the costly manual writing of explanation annotation from scratch, greatly speeding up our data annotation process.
\section{Experiments}
\label{sec:experiments}

\subsection{Data Analysis}

We have collected $280$ hours of driving data for training. These data consist of 50k routes, collected in 30 driving scenarios with different weather and lighting conditions across 8 maps (Town01, Town02, Town03, Town04, Town06, Town07, Town10HD, Town12) in CARLA. On average, each scenario has about 200 trigger points on each map to be randomly triggered. Each scenario is either a common or rare safety-critical situation in driving. For each frame, we collect images from 4 cameras on the front, rear, left, and right, and also the point clouds from a LiDAR sensor added in the center of the ego vehicle. All data  has corresponding explanations and accurate decisions that successfully drive through scenarios. 

\tabref{tab:data_comparison} presents the comparison with the previous datasets designed for driving understanding with natural language. Our data has two unique features. The first is the alignment of behavioral planning states. This enables us to transform the MLLM planner's output to control signal so that our framework can control vehicles in closed-loop driving. The second is human interaction annotation. It is characterized by natural language instructions given by humans alongside the responding decisions and explanations. The objective is to improve the ability to understand human instructions and respond accordingly.

\begin{table}[t]
  \centering
	\renewcommand{\arraystretch}{1.1}
	\setlength{\tabcolsep}{0.8mm}
  \begin{tabular}{lccccc} \hline
    Dataset & Perception & Reason & Plan & Align & Interact \\ \hline
    NuPrompt~\cite{wu2023language} & $\checkmark$ &  &  &  &  \\
    NuScenes-QA~\cite{qian2023nuscenes} & $\checkmark$ & $\checkmark$ &  &  &  \\
    Rank2Tell~\cite{sachdeva2023rank2tell} & $\checkmark$ & $\checkmark$ &  &  &  \\
    BDD-X~\cite{xu2023drivegpt4} &  & $\checkmark$ & $\checkmark$ &  &  \\
    DRAMA~\cite{movva2023large} &  & $\checkmark$ & $\checkmark$ &  &  \\
    DriveLM~\cite{drivelm2023} & $\checkmark$ & $\checkmark$ & $\checkmark$ &  &  \\
    Ours & $\checkmark$ & $\checkmark$ & $\checkmark$ & $\checkmark$ & $\checkmark$ \\ \hline
  \end{tabular}
  \caption{\textbf{Comparisons of AD datasets for driving understanding}. The alignment of behavioral planning states enables us to transform the MLLM planner's output to control signal for closed-loop driving. The human interaction annotation enhances the model's understanding of customized language instruction. }
  \label{tab:data_comparison}       
\end{table}

Our training data contains 30 common or rare safety-critical scenarios, and \tabref{tab:scenario_list} lists the names of all the scenarios and describes the source of the scenarios. Non-custom scenarios (marked as $\dagger$ and $\ddagger$) are usually set by loading preset trigger points, which makes it difficult to set them in other maps. Therefore, we have dynamized the scenarios to automatically find suitable trigger points on any map in preparation for the scenario setup. It is worth noting that all scenarios in  \tabref{tab:scenario_list} have been dynamized.

\begin{table}[t]
  \centering
  \renewcommand{\arraystretch}{1.1}
  \setlength{\tabcolsep}{6mm}
  \begin{tabular}{c|l} \hline
    ID & Scenario Name \\ \hline
     1 & YieldBehindEmergencyVehicles ($\star$) \\
     2 & OvertakingFromLeft ($\star$) \\
     3 & OvertakingFromRight ($\star$) \\
     4 & LeftBorrowPassObstacle ($\dagger$) \\
     5 & LeftBorrowPassAccident ($\dagger$) \\
     6 & LeftInvasionBorrowPassObstacle ($\dagger$) \\
     7 & LeftInvasionBorrowPassAccident ($\dagger$) \\
     8 & RightBorrowPassObstacle ($\dagger$) \\
     9 & RightBorrowPassAccident ($\dagger$) \\
    10 & RightInvasionBorrowPassObstacle ($\dagger$) \\
    11 & RightInvasionBorrowPassAccident ($\dagger$) \\
    12 & JunctionRightChange ($\star$) \\
    13 & JunctionLeftChange ($\star$) \\
    14 & JunctionStraight ($\star$) \\
    15 & JunctionYieldPedestrian ($\star$) \\
    16 & JunctionYieldPedestrianAfterTurn ($\dagger$) \\
    17 & YieldJunctionSpecialisedVehicles ($\dagger$) \\
    18 & LeftChangeInRoute ($\star$) \\
    19 & RightChangeInRoute ($\star$) \\
    20 & UnprotectedJunctionLeftTurn ($\dagger$) \\
    21 & UnprotectedJunctionStraight ($\dagger$) \\
    22 & UnprotectedJunctionRightTurn ($\dagger$) \\
    23 & SignedJunctionLeftTurn ($\dagger$) \\
    24 & SignedJunctionStraight ($\dagger$) \\
    25 & SignedJunctionRightTurn ($\dagger$) \\
    26 & PedestrianBlindSpotA ($\ddagger$) \\
    27 & PedestrianBlindSpotB ($\ddagger$) \\
    28 & VehicleBlindSpotA ($\ddagger$) \\
    29 & VehicleBlindSpotB ($\ddagger$) \\
    30 & FollowerChange ($\dagger$) \\ \hline
  \end{tabular}
    \caption{\textbf{Scenario list.} $\star$ denotes that these scenarios are constructed by ourselves. $\dagger$ and denotes that these scenarios are from official Carla settings and ReasonNet\cite{shao2023reasonnet}, respectively.}
  \label{tab:scenario_list}
\end{table}

\begin{table*}[t]
  \centering
  \renewcommand{\arraystretch}{1.3}
  \setlength{\tabcolsep}{1mm}
  \begin{tabular}{lccccccccccccc} \hline
    \multirow{2}{*}{Method} & \multirow{2}{*}{Type} & \multirow{2}{*}{Acc. $(\%)$ $\uparrow$} & 
    \multicolumn{3}{c}{Path (F1) $\uparrow$} & \multicolumn{4}{c}{Speed (F1) $\uparrow$} & 
    \multirow{2}{*}{BLEU-4 $\uparrow$} & \multirow{2}{*}{CIDEr $\uparrow$} & \multirow{2}{*}{METEOR $\uparrow$} \\ 
    & & & follow & change & borrow & keep & accel. & decel. & stop & & & \\ \hline

    LLaVA 1.5~\cite{liu2023improved} & LLM & 22.92 & 0.73 & 0.00 & 0.00 & 0.75 & 0.00 & 0.02 & 0.00 & 10.00 & 18.03 & 23.00 \\
    InstructBLIP~\cite{instructblip} & LLM & 17.92 & 0.00 & 0.30 & 0.08 & 0.23 & 0.00 & 0.28 & 0.00 & 9.81 & 18.61 & 22.95 \\
    Apollo~\cite{apollo_auto} & FSM & 18.53 & 0.76 & 0.40 & 0.04 & 0.54 & 0.05 & 0.19 & 0.37 & - & - & - \\ \hline
    DriveMLM & LLM & \textbf{75.23} & \textbf{0.90} & \textbf{0.52} & \textbf{0.89} & \textbf{0.91} & \textbf{0.61} & \textbf{0.66} & \textbf{0.89} & \textbf{40.46} & \textbf{124.91} & \textbf{56.54} \\ \hline
  \end{tabular}
  \caption{\textbf{Results of open-loop evaluation on CARLA Town05.} Compared with previous approaches, our method can predict more precise decisions and give better explanations for the decision choice.}
  \label{tab:text_metric}
\end{table*}

\subsection{Implementation Details} 

Our MLLM model is built from LLaMA~\cite{gao2023llama}. Specifically, we use ViT-g/14 from EVA-CLIP~\cite{fang2023eva} as the visual encoder and LLaMA-7B \cite{touvron2023llama} as the LLM. The querying transformer with $N_Q$ queries is applied to extract image tokens from ViT, where we set $N_Q=32$. For the LiDAR encoder, we use the GD-MAE~\cite{yang2023gd} model finetuned on ONCE~\cite{mao2021one}. Based on the pre-trained husky model, we train MLLM with instruction following data.
We employ the AdamW optimizer with $\beta_1=0.9$, $\beta_2=0.95$, and a cosine learning rate decay with learning rate $5e^{-5}$. The training epoch is 2, and the batch size is 256.
We train QFormer and LLM to ensure the instruction following ability of LLM so that we can obtain a predefined format of path decision and speed decision. The resolution of image input to MLLM is set as $448\times448$.

For evaluating closed-loop driving performance, we use the widely used Town05Long benchmark, which follows previous work~\cite{chitta2022transfuser, shao2023safety}. It is worth noting that Town05 is not in our training data. We use Driving Score (DS), Route Completion (RC), and Infraction Score (IS)~\cite{dosovitskiy2017carla} as the metrics. RC computes the average percentage of routes completed by an agent. IS measures the infraction penalty between $0$ and $1$, including collision and violation of traffic rules. Note that IS is only calculated on the completed part of a route. DS is the core metric among the three, which is the product of both RC and IS. We also evaluate driving performance using Miles Per Intervention (MPI), which is a widely used metric in industry.It is computed as the total distance traveled over the total times of human takeovers. If the ego-car violates traffic rules or has a collision, it will be taken over and continue self-driving in a safe location until it reaches its destination. Unlike DS, which terminates the route under certain conditions, MPI requires the ego-car to complete the entire route. 

For the open-loop evaluation, we collect 10 routes of each scenario in Town05 obtained and annotated by human drivers as the test set. To evaluate decision prediction, we compute the accuracy of predicted decision pairs and the F1 score of each type of decision.

For the explanation prediction task, we use the commonly used metrics in the NLP community, including BLEU-4~\cite{papineni2002bleu}, CIDEr~\cite{vedantam2015cider} and METEOR~\cite{banerjee2005meteor}. 
We compare our method with the popular Apollo, which is based on a finite state machine (FSM) and two MLLM models - LLaVA1.5~\cite{liu2023improved} and InstructBLIP~\cite{instructblip}.
These two MLLM models used for comparison were not fine-tuned but instead provided with several examples of input/decision pairs for few-shot adaptation. 

\subsection{Evaluation of Driving Knowledge}

We adopt open-loop evaluation to evaluate the driving knowledge, which includes the decision prediction and the explanation prediction task. \tabref{tab:text_metric} presents the accuracy of predicted decision pairs, F1-score of each type of decision for the decision prediction, and BLEU-4~\cite{papineni2002bleu}, CIDEr~\cite{vedantam2015cider} and METEOR~\cite{banerjee2005meteor} for the predicted explanation. For Apollo, manually collected scenarios on Town05 are replayed as input to models in \tabref{tab:text_metric}. The corresponding model states and outputs at every timestamp of replay are saved as predictions for metric calculation. For other methods, we give them the corresponding images as input and the proper prompts.
By comparing model prediction with our manually collected ground truth, accuracy reveals decision correctness and similarity to human behavior, and the F1-score demonstrates the decision-making capability across each individual type of path and speed decision. DriveMLM achieves the highest accuracy overall, surpassing LLaVA with an accuracy of 40.97\%. Compared to the Apollo baseline, the higher F1-score 
of DriveMLM suggests that it is much more effective in overtaking the rule-based state machine for solving various road situations. LLaVA~\cite{liu2023improved}, InstructBLIP~\cite{instructblip}, and our proposed DriveMLM can output explanations of decisions in the form of question and answer. In terms of BLEU-4, CIDEr, and METEOR, DriveMLM can achieve the highest performance, indicating that DriveMLM can give the most reasonable explanation of the decision.

\subsection{Evaluation in Closed-Loop Driving}

We evaluate closed-loop driving in CARLA, the most widely used and realistic simulation benchmark publicly available. State-of-the-art methods~\cite{zhang2021end, shao2023safety, jia2023think} that are capable of performing closed-loop driving in CARLA are included for performance comparison. The open-sourced Apollo~\cite{apollo_auto} is also evaluated in CARLA as a baseline. No other LLM-based methods have shown the readiness to be deployed and evaluated besides ours. All methods are evaluated on Town05 long benchmarks~\cite{chitta2022transfuser}. 

\tabref{tab:closeloop_metric} presents the Driving Score, Route Completion, and Infraction Score. Note that despite being a rule-based method, Apollo achieves almost on-par performance with recent end-to-end methods. DriveMLM surpasses all other methods on Driving Score by a large margin.This suggests that DriveMLM is better for handling state-transitions to safely drive through hard cases. 
The last column in \tabref{tab:closeloop_metric} presents the results of MPI evaluation. This metric shows a more holistic driving performance because an agent is required to finish all routes. In other words, all situations along all routes are encountered by the tested agents. Thinktwice achieves better DS but lower MPI than Interfuser due to frequently crossing the stop line. However, CARLA imposes minimal penalties for this behavior. By contrast, MPI takes each violation of traffic rules as one take-over. DriveMLM also achieves the highest MPI among all other methods, suggesting its ability to avoid more situations for a safer driving experience.

\begin{table}[t]
  \centering
	\renewcommand{\arraystretch}{1.1}
	\setlength{\tabcolsep}{2mm}
  \begin{tabular}{lccccc} \hline
    Method & Type & DS$\uparrow$ & RC$\uparrow$ & IS$\uparrow$ & MPI$\uparrow$ \\ \hline
    Roach~\cite{zhang2021end} & DD & 43.6 & 80.4 & 0.54 & - \\
    Interfuser~\cite{shao2023safety} & DD & 68.3 & 95.0 & 0.72 & 0.70 \\
    ThinkTwice~\cite{jia2023think} & DD & 70.9 & 95.5 & 0.75 & 0.40 \\
    Apollo~\cite{apollo_auto} & FSM & 71.4 & 92.2 & \textbf{0.80} & 0.76 \\ \hline
    DriveMLM & LLM & \textbf{76.1} & \textbf{98.1} & 0.78 & \textbf{0.96} \\ \hline
  \end{tabular}
  \caption{\textbf{Results of closed-loop driving on CARLA Town05 Long.} FSM denotes a Finite State Machine. DD denotes Data Driven. DS denotes Driving Score. RC denotes Route Completion. IS denotes Infraction Score. MPI denotes Miles Per Intervention. DriveMLM has a higher driving score and route completion rate and is also close to Apollo's infraction penalty, indicating that DriveMLM can make better decisions while following the traffic rules. Meanwhile, DriveMLM also shows advantages in MPI, representing fewer human takeovers at the same mileage.}
  \label{tab:closeloop_metric}       
\end{table}

\subsection{Ablation Study}

\textbf{Sensor Modality.} \tabref{tab:ablation_modal} presents the results of different impacts of input sensor modality to the DriveMLM. Multi-View (MV) images bring a substantial performance improvement in both path and speed F1-score, along with 18.19\% increase in accuracy. Compared to concatenating temporal tokens directly, temporal QFormer results in a larger improvement of 7.4\%, while ensuring multi-modal decision capability, which leads to 0.05 improvement in the average F1-score on speed decision. Point clouds do not show the ability to enhance performance.

\textbf{Temporal Module Design.} 
We propose the temporal QFormer module to process the temporal multi-view images. A simple and naive design is directly concatenating query tokens temporal to generate $N_{tq} = T\times N_I \times N_Q$ tokens acting as LLM input. But $N_{tq}$ increases with $T$, contributing to large computational costs. Instead, we propose the temporal QFormer module to process temporal images for each view separately, generating $N_I \times N_Q$ tokens for LLM input. The comparison of the temporal module is shown in table \ref{tab:ablation_modal}, indicating the better performance of our temporal module design with fewer image tokens. We set $T=2$ by default in our experiments.

\begin{table}[t]
  \centering
  \renewcommand{\arraystretch}{1.1}
  \setlength{\tabcolsep}{1mm}
  \begin{tabular}{ccccccc} \hline
    MV & CT & TQ & PC & Acc. $(\%)$ $\uparrow$ & Path (F1) $\uparrow$ & Speed (F1) $\uparrow$ \\ \hline
    - & - & - & - & 47.83 & 0.55 & 0.61 \\
    \checkmark & - & - & - & 64.54 & \textbf{0.78} & 0.70 \\
    \checkmark & \checkmark & - & - & 67.22 & 0.70 & 0.68 \\
    \checkmark & - & \checkmark & - & \textbf{75.23} & \textbf{0.78} & \textbf{0.75} \\
    \checkmark & - & \checkmark & \checkmark & 74.99 & 0.77 & \textbf{0.75} \\ \hline
  \end{tabular}
  \caption{\textbf{Ablation results of sensor modality and temporal information.} MV denotes multi-view images, CT denotes concatenating temporal tokens, TQ denotes temporal QFormer, and PC denotes point clouds. MV + TQ shows the best decision performance, and CT brings a small improvement in accuracy but leads to greater computational consumption. PC has little impact on DriveMLM. This might be caused by the large representation gap between the sparse pyramid transformer and the MLLM decoder.}
  \label{tab:ablation_modal}
\end{table}

\subsection{Case Study \& Visualization}

\begin{figure}[t]
  \includegraphics[width=1.0\linewidth]{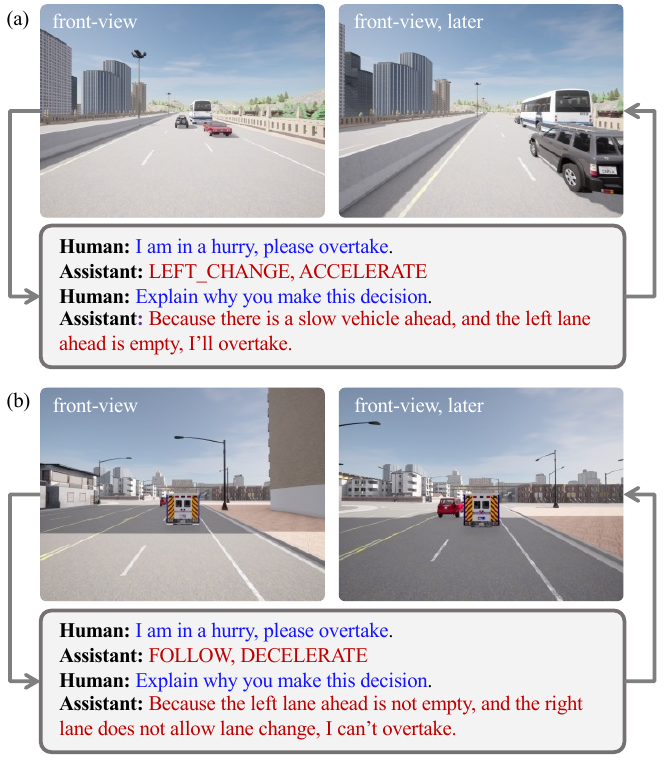}
  \caption{\textbf{Visualization of human interaction with DriveMLM}, featuring examples of (a) a successful overtake and (b) a failed overtake due to unreasonable demands rejected by DriveMLM. In each example, the left image captures the moment of giving instructions, while the right image represents a subsequent stage of instruction execution. }
  \label{fig:vis_interact}   
\end{figure}

\textbf{Human Interaction.} \figref{fig:vis_interact} provides an example of how vehicle control can be achieved through human instructions. 
The control process involves analyzing the road conditions, making decision choices, and providing explanatory statements. When given the identical instruction to ``overtake", DriveMLM exhibits varying responses based on the analysis of the current traffic conditions. In the scenario where the right lane is occupied and the left lane is available, the system opts to overtake from the left. 
However, in situations where the given instruction may pose a danger, such as when all lanes are occupied, DriveMLM chooses to refrain from executing the overtaking maneuver and responds appropriately.
DriveMLM, in this context, serves as an interface for human-vehicle interaction, which evaluates the reasonableness of the instruction based on traffic dynamics and ensures its compliance with predefined rules before ultimately selecting a course of action.

\begin{figure*}[t]
  \centering
  \subfigure[Unknown objects blocking the path]{
    \includegraphics[width=0.48\linewidth]{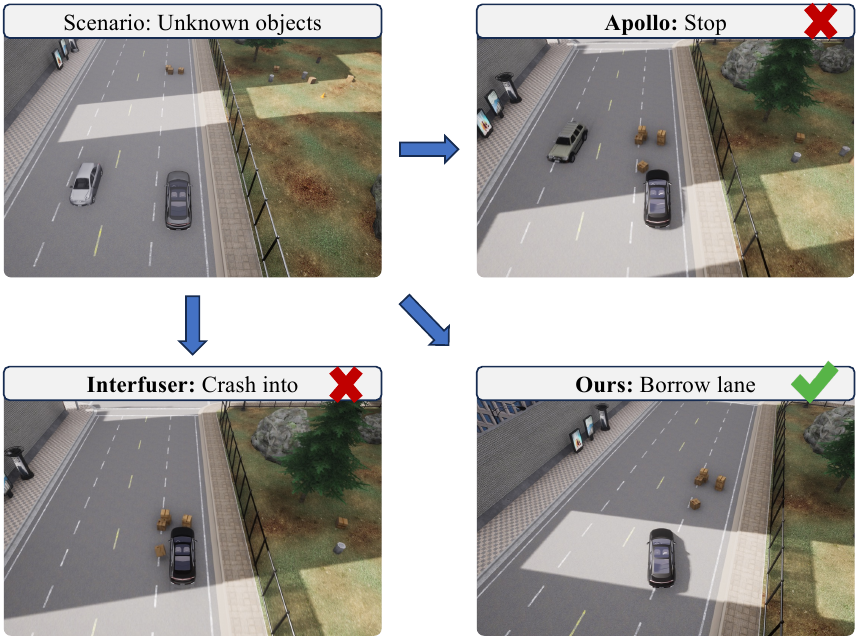}
    \label{fig:vis_model_comp_a}
  }%
  \hfill
  \subfigure[Emergency vehicle approaching from behind]{
    \includegraphics[width=0.48\linewidth]{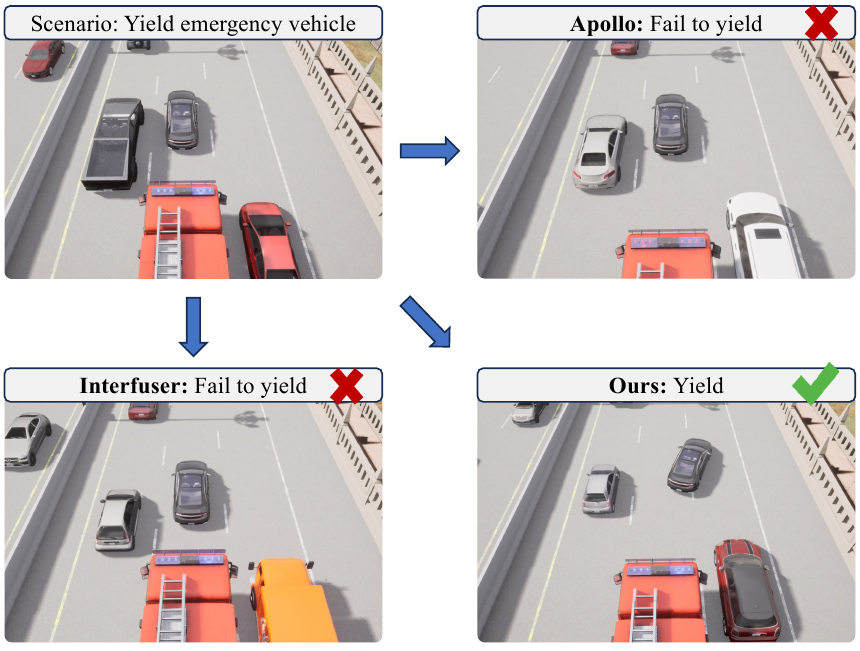}
    \label{fig:vis_model_comp_b}
  }
    \caption{\textbf{Comparison between Interfuser, Apollo, and DriveMLM when confronted with unknown objects (a) and an emergency vehicle behind (b).} (a) Interfuser ignores the obstacles and causes a crash, Apollo stops before the unknown objects, and DriveMLM successfully bypasses the obstacles by a borrow lane decision. (b) Interfuser and Apollo do not yield to the firetruck behind, but DriveMLM successfully changes lanes to the right and yields to the firetruck.}
  \label{fig:vis_model_comp}
\end{figure*}

\textbf{Comparative Analysis of Diverse Methods.}
Compared to methods such as Interfuser~\cite{shao2023safety} or Apollo~\cite{apollo_auto}, our approach demonstrates superior performance in scenarios with unknown obstacles or those necessitating common sense. As depicted in \figref{fig:vis_model_comp} (a), when facing unknown obstacles on the road, previous methods typically either overlook them or halt the vehicle, both strategies deviating from optimal driving practices. In contrast, our method employs a more logical 'borrow lane' decision, effectively preventing accidents. In addition, the deficiency of previous methods in embodying real-world common sense or understanding traffic rules limits their capability to manage diverse special scenarios encountered in complex driving scenarios. Illustratively, as shown in \figref{fig:vis_model_comp} (b), when emergency vehicles approaching from behind, conventional methods fail to yield, whereas our method proactively clears the path for the firetruck.

\begin{figure}[t]
  \includegraphics[width=1.0\linewidth]{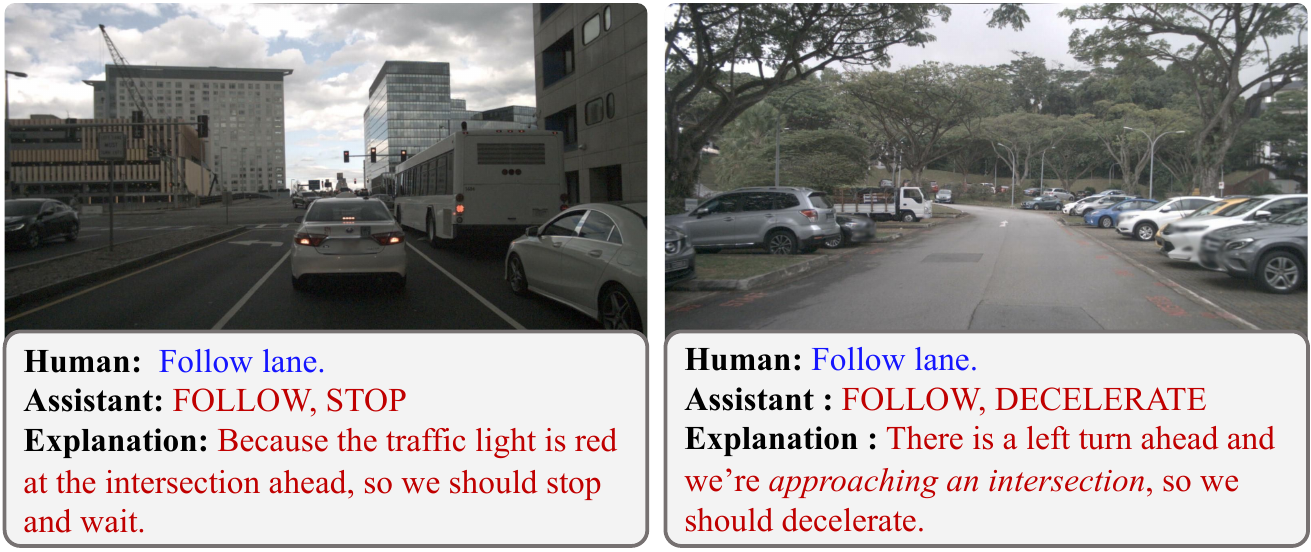}
  \caption{\textbf{The zero-shot performance of DriveMLM on the real driving scene from nuScenes dataset.} (a) DriveMLM can recognize the red light and stop (b) DriveMLM can infer the location of the intersection and slow down in advance.}
  \label{fig:nuscene}   
\end{figure}

\textbf{Performance in Real Scenarios.} 
We evaluate DriveMLM on the nuScenes dataset \cite{caesar2020nuscenes} to test the zero-shot performance of the developed driving system.
We annotate 6,019 frames on the validation set, and the zero-shot performance of decision accuracy is 0.395.
Fig.~\ref{fig:nuscene} presents the result on two real driving scenes, indicating the generability of DriveMLM.

\begin{table}[t]
  \centering
  \renewcommand{\arraystretch}{1.1}
  \setlength{\tabcolsep}{2.0mm}
  \begin{tabular}{lccccc} \hline
    Model & \multicolumn{2}{c}{Lmdeploy} & \multicolumn{2}{c}{vLLM} \\ 
    & 1 img & 4 imgs & 1 img & 4 imgs \\ \hline
    LLaVA-1.5 & 0.0111 & 0.0129 & 0.0151 & 0.0158 \\
    mini-InternVL & 0.0037 & 0.0039 & / & / \\
    F.t. LLaVA-1.5 & 0.0122 & 0.0138 & 0.0163 & 0.0169 \\
    F.t. LLaVA-1.5+Lidar & 0.0129 & 0.0142 & 0.0168 & 0.0176 \\ \hline
  \end{tabular}
  \caption{\textbf{Inference Speed Results of Different Model Sizes under Different Optimization Frameworks (lmdeploy and vLLM)}.  F.t. denotes the fine-tuned model.}
  \label{tab:inference_speed}
\end{table}

\textbf{Model Inference speed.} 
In autonomous driving systems, model inference speed is crucial. The model needs to make quick decisions in the current scenario; otherwise, prolonged latency can lead to changes in the car’s state and the environment, causing decision failures or delays. Therefore, we tested the inference speed (s/token) of different models, as shown in \tabref{tab:inference_speed}. We tested the inference speed of state-of-the-art open-source models like LLaVA-1.5-7b and mini-InternVL-2b, and also tested the inference speed of fine-tuned LLaVA and fine-tuned LLaVA with Lidar input. We found that the speed of mini-InternVL-2b, with reduced model parameters, decreased to about 2b parameters, indicating that lightweight models may have the potential to enable real-time decision-making. We compared lmdeploy and vLLM methods for accelerating model inference. And We found that the inference speed of the model optimized with lmdeploy is superior to that optimized with vLLM. Therefore, optimizing and accelerating the network structure is also a potential approach to achieving real-time decision-making.

\section{Conclusion}
\label{sec:conclusion}

In this work, we have presented DriveMLM, a novel framework that leverages large language models (LLMs) for autonomous driving (AD). DriveMLM can perform close-loop AD in realistic simulators by using a multi-modal LLM (MLLM) to model the behavior planning module of a modular AD system. DriveMLM can also generate natural language explanations for its driving decisions, which can enhance the transparency and trustworthiness of the AD system. We have shown that DriveMLM can outperform the Apollo baseline on the CARLA Town05 Long benchmark. We believe that our work can inspire more research on the integration of LLMs and AD.

\begin{small}
\section*{Declarations}

\vspace{.3in} \noindent \textbf{Abbreviations}\\
AD: Autonomous Driving; LLM: large language model; MLLM: multi-modal large language model; DS: Driving Score; MPI: Miles Per Intervention, BEV: Bird's-Eye-View; SST: Single-stride Sparse Transformer; MV: Multi-View;

\vspace{.3in} \noindent \textbf{Data Availability}\\
The datasets generated during and analyzed during the current study are available from the corresponding author on reasonable request.

\vspace{.3in} \noindent \textbf{Funding}\\
Not applicable

\vspace{.3in} \noindent \textbf{Competing Interests}\\
Wenhai Wang is an Associate Editor at Visual Intelligence and was not involved in the editorial review of this article or the decision to publish it. The authors declare that they have no other competing interests.

\vspace{.3in}\noindent \textbf{Authors’ Contributions}\\  
All authors contributed to the study’s conception and design. 
Conceptualization: WhW, ZQL, GL, EC; 
Methodology: WhW, ZQL, JWX, EC; 
Formal analysis and investigation: JWX, HmZ, HmD, EC; 
Writing – original draft preparation: WhW, YW, EC; 
Writing – review and editing: WhW, SLW, JWX; 
Resources: JD, LL, GL; 
Supervision: JD, XZZ. 
All authors read and approved the final manuscript.

\vspace{.3in} \noindent \textbf{Author details}\\
$^{1}$Shanghai Jiao Tong University, Shanghai, 200240, China.
$^{2}$Shanghai Artificial Intelligence Laboratory, Shanghai, 200232, China.
$^{3}$The Chinese University of Hong Kong, Hong Kong, 999077, China.
$^{4}$SenseTime Research, Shanghai, 200233, China.
$^{5}$Stanford University, Stanford, CA 94305, United States.
$^{6}$Nanjing University, Nanjing, 210023, China.
$^{7}$Tsinghua University, Beijing, 100084, China.
$^{8}$Beijing National Research Center for Information Science and Technology, Beijing, 100084, China.

\end{small}


\bibliographystyle{unsrt}
\bibliography{reference}

\end{document}